\documentclass[sigconf]{acmart}

\usepackage{amsmath}
\usepackage{booktabs}
\usepackage{tabularx}
\usepackage{adjustbox}
\usepackage{url}
\usepackage{mdwlist}
\usepackage{amsfonts}
\usepackage{multirow}
\usepackage{array}
\usepackage{epstopdf}
\usepackage{enumitem}
\usepackage[switch]{lineno}
\usepackage{subfigure}
\usepackage{bm}
\usepackage{dcolumn}

\AtBeginDocument{%
  \providecommand\BibTeX{{%
    \normalfont B\kern-0.5em{\scshape i\kern-0.25em b}\kern-0.8em\TeX}}}

\copyrightyear{2024}
\acmYear{2024}
\setcopyright{acmlicensed}\acmConference[WWW '24 Companion]{Companion Proceedings of the ACM Web Conference 2024}{May 13--17, 2024}{Singapore, Singapore}
\acmBooktitle{Companion Proceedings of the ACM Web Conference 2024 (WWW '24 Companion), May 13--17, 2024, Singapore, Singapore}
\acmDOI{10.1145/3589335.3648330}
\acmISBN{979-8-4007-0172-6/24/05}

\begin{document}

\title{FinReport: Explainable Stock Earnings Forecasting via \\ News Factor Analyzing Model}

\author{Xiangyu Li}
\authornote{Both authors contributed equally to the paper}
\email{65603605lxy@gmail.com}
\orcid{0009-0002-8261475r}
\affiliation{
  \institution{South China University of Technology}
  \country{}
}
\author{Xinjie Shen}
\authornotemark[1]
\email{frinkleko@gmail.com}
\orcid{0009-0004-9176-5400}
\affiliation{%
  \institution{South China University of Technology}
  \country{}
}

\author{Yawen Zeng}
\orcid{0000-0003-1908-1157}
\email{yawenzeng11@gmail.com}
\affiliation{
  \institution{ByteDance AI Lab}
  \country{}
}

\author{Xiaofen Xing}
\authornote{Corresponding authors}
\email{xfxing@scut.edu.cn}
\orcid{0000-0002-0016-9055}
\affiliation{
  \institution{South China University of Technology}
  \country{}
}
\author{Jin Xu}
\authornotemark[2]
\email{jinxu@scut.edu.cn}
\orcid{0009-0001-8735-3532}
\affiliation{
  \institution{South China University of Technology\\ Pazhou Lab}
  \country{}
}

\renewcommand{\shortauthors}{Li, et al.}

\begin{abstract}
The task of stock earnings forecasting has received considerable attention due to the demand investors in real-world scenarios. However, compared with financial institutions, it is not easy for ordinary investors to mine factors and analyze news. On the other hand, although large language models in the financial field can serve users in the form of dialogue robots, it still requires users to have financial knowledge to ask reasonable questions. To serve the user experience, we aim to build an automatic system, FinReport, for ordinary investors to collect information, analyze it, and generate reports after summarizing. 

Specifically, our FinReport is based on financial news announcements and a multi-factor model to ensure the professionalism of the report. The FinReport consists of three modules: news factorization module, return forecasting module, risk assessment module. The news factorization module involves understanding news information and combining it with stock factors, the return forecasting module aim to analysis the impact of news on market sentiment, and the risk assessment module is adopted to control investment risk. Extensive experiments on real-world datasets have well verified the effectiveness and explainability of our proposed FinReport. Our codes and datasets are available at https://github.com/frinkleko/FinReport.

\begin{figure}[h]
    \center
    \includegraphics[width=0.49\textwidth]{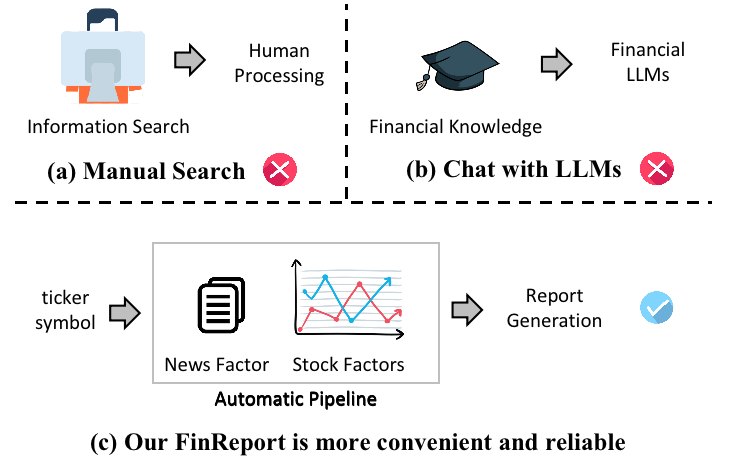}
    \vspace{-0.2cm}
    \caption{Examples of the manual search, Chat with LLMs and our FinReport solution.}
    \label{fig:example}
    \vspace{-0.5cm}
\end{figure}

\begin{figure*}[t]
    \centering
    \includegraphics[width=0.95\textwidth]{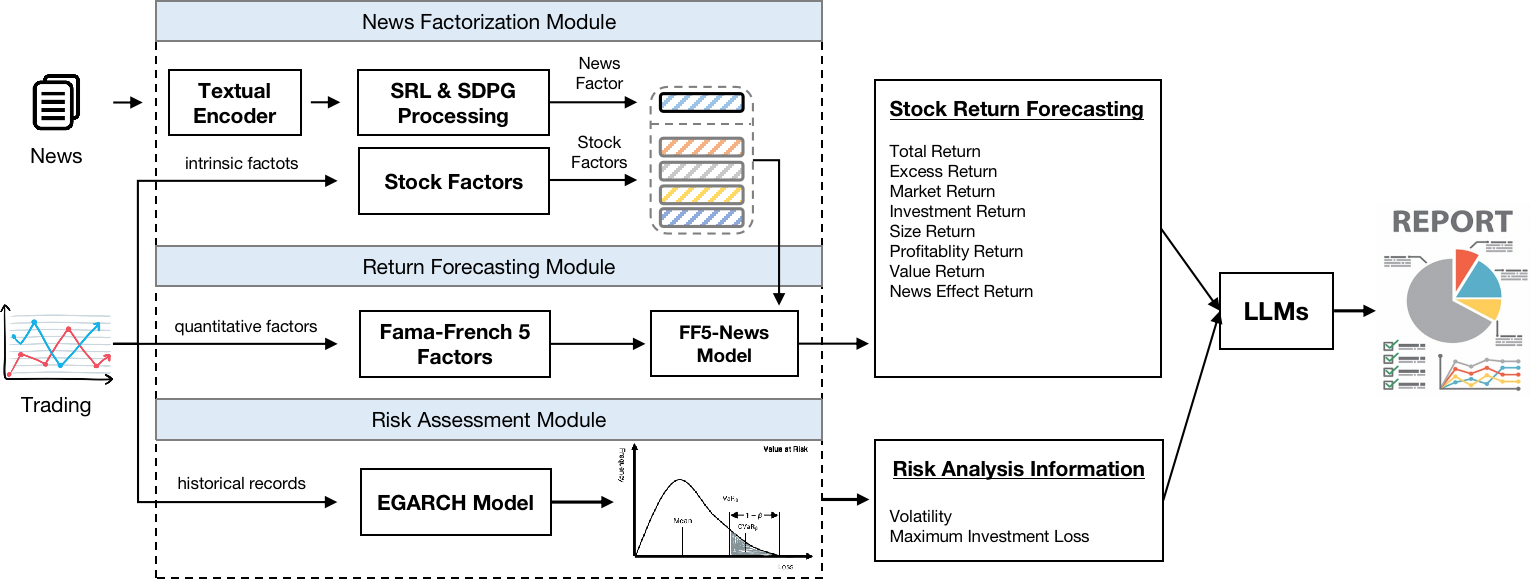}
        \vspace{-0.3cm}
    \caption{The overall architecture of our FinReport for explainable stock returns forecasting.}
    \label{fig:framework}
    \vspace{-0.45cm}
\end{figure*}

\end{abstract}





\maketitle

\section{Introduction}
The endeavor to forecast the ever-changing stock market has always been a captivating task for investors. In particular, its randomness, volatility, and behavioral diversity of participants have posed significant challenges. Among them, stock data and news data are crucial factors that influence investors' decision-making. 1) Stock data refers to numerical data that characterizes stocks in time series. For instance, \cite{CHAUDHARI2023103293} proposed a novel data quantization and fusion way of stock data for stock time series predictions.
2) News data is unstructured text that contains complex time-sensitive information. Pioneers are utilizing natural language processing (NLP) tools\cite{zeng2022keyword,zeng2021multi,zeng2022prompt} to comprehend news and facilitate stock forecasting \cite{peng2024energybased,zeng2024cote}. For example, \cite{Cai2018} exacted key word by TF-IDF and contacted it with index.  ~\citet{Sawhney2020} focused on social media text and correlations among stocks, proposed a hierarchical temporal fusion to process index and news data.

While the field of stock forecasting is thriving, unfortunately not everyone can benefit from it. In fact, well-known investment institutions can build complex and comprehensive systems of quantitative and public opinion analysis to sensitively detect market changes. However, this is not an easy task for the majority of ordinary investors. As shown in Fig. \ref{fig:example}(a), where manually collecting news, reading financial reports, and building small factor analysis models are common strategies that are tedious and unstable. On the other hand, with the emergence of ChatGPT \cite{pan2023llms}, large language models (LLMs) in the financial field like BloombergGPT can serve users in the form of dialogue robots, as shown in Fig. \ref{fig:example}(b). However, this requires users to have financial knowledge to ask reasonable questions. Therefore, in this paper, we aim to build an automatic system for ordinary investors to collect information, analyze it, and generate reports after summarizing. As shown in Fig. \ref{fig:example}(c), the user only needs to input the ticker symbol or news, and our system has the ability to generate a customized report (e.g. stock returns forecasts) for the user, which is more convenient and reliable.

To obtain a comprehensive report, we focus on two main components: stock data and news data. Previous studies have attempted to directly combine news and stock factors, disregarding the fundamental differences between the continuity and density of stocks and the discontinuity and sparseness of news. Moreover, the impact of news on the stock market is subject to "chronological deviation," meaning that news events may produce varying reactions at different timestamps. Therefore, capturing the dynamic relationship between news events and the market requires: 1) News factorization, which involves understanding news information and combining it with stock factors. 2) Return forecasting, which includes assessing the impact of news on market sentiment and the impact of news announcements on a company's stock. 3) Risk assessment, which is crucial for users to effectively participate in financial markets. Ultimately, our system will provide users with effective investment recommendations based on these elements.

To address the aforementioned challenges, we propose FinReport, an explainable model for reporting stock returns. Specifically, our FinReport is based on financial news announcements and a multi-factor model to ensure the professionalism of the report. The FinReport consists of three modules: 1) News factorization module, which combines Semantic role labeling and semantic dependency parsing graph to comprehend news information and then combines it with stock multi-factors to predict news announcement classification. 2) Return forecasting module, which utilizes the Fama-French 5-factor model to analyze the sentiment impact of the news on the market. 3) Risk assessment module, which employs the EGARCH model to build a VaR risk assessment for the stock risk level based on historical fluctuation information. Finally, the above information is fed into our LLM to generate a readable report. Extensive experiments demonstrate that our model achieves a higher ROI and Sharpe Ratio. The main contributions are summarized as follows:

\begin{itemize}[leftmargin=*]
\item To the best of our knowledge, this is the first work that introduces a financial report model to automatically collects information, analyzes, and summarizes.
\item We propose three sub-modules to respectively address news factorization, return
forecasting, and risk assessment, which make reporting more reliable.
\item Extensive experiments conducted on real-world datasets demonstrate the effectiveness and explainability of our solution.
\end{itemize}

\section{Related Work}
\label{sec:relwo}
\subsection{Quantitative Finance}
  Quantitative finance is an applied mathematics field that focuses on financial markets, which involves the use of mathematical and statistical methods to analyze financial markets \cite{kou2019machine,kanamura2021pricing}.  The field of quantitative finance can be divided into two main branches: derivatives pricing\cite{tavella2003quantitative,horvath2021deep} and risk analysis\cite{mieg2022volatility}. Derivatives pricing is concerned with the pricing of financial instruments, such as options, while risk analysis focuses on models and techniques for measuring and managing the risk of complex financial instruments, such as credit derivatives.In order to quantify the risk of financial investments more accurately, the VaR (Value at Risk) \cite{VillarRubio2023} method is commonly employed. It estimates the maximum loss that an asset or investment portfolio may suffer over a given period of time. 

\subsection{News Semantic Understanding}
In the realm of news semantic understanding, pivotal roles are played by Semantic Role Labeling (SRL) \cite{marquez2008semantic} and Semantic Dependency Parsing Graph (SDPG) \cite{agic2015semantic} techniques. SRL aims to identify predicates within sentences and label associated semantic roles, like agents and patients. In parallel, SDPG transforms sentences into a structured form represented as directed graphs, where arcs link pairs of words. Since the introduction of standard datasets\cite{Oepen2015}, numerous SDPG methods have been explored on the basis of syntactic analysis\cite{Dozat2018, Peng2018, Oepen2020,Wang2020}. Decoding methods for SDPG include integer linear programming\cite{Almeida2015} and transition-based approaches introducing novel shift-reduce automata\cite{Zhang2016}. By employing semantic understanding techniques, we can extract semantic roles and semantic dependency relationships within sentence components, enabling more detailed and profound semantic analysis.

\section{Proposed Method}\label{sec:method}
In this paper, we try to automatically generate a comprehensive report including news analysis and quantitative understanding, so as to lower the threshold for ordinary users. Towards this goal, our FinReport consists of three components, as shown in Fig. \ref{fig:framework}: news factorization module, return forecasting module, and risk assessment module. Finally, based on the return forecasting and risk assessment results obtained from the aforementioned components, we will automatically generate readable and explainable reports with the help of LLMs.

\subsection{Preliminaries}\label{subsec:pre}
The stock return forecasting task is defined as predicting future returns based on historical data. Stock data includes opening price, closing price, highest price, lowest price and trading volume, with related news. Among them, the news data is mainly the news title with the stock name, which can be ``None'', because the news does not happen every day.



\subsection{News Factorization Module}

Financial news and stock factors are vital parts of market info, crucial for report quality. In this section, we factorize news by extracting information and merging it with quantitative factors, which are used to predict return classifications.

\begin{figure}[t]
    \centering
    \includegraphics[width=0.45\textwidth]{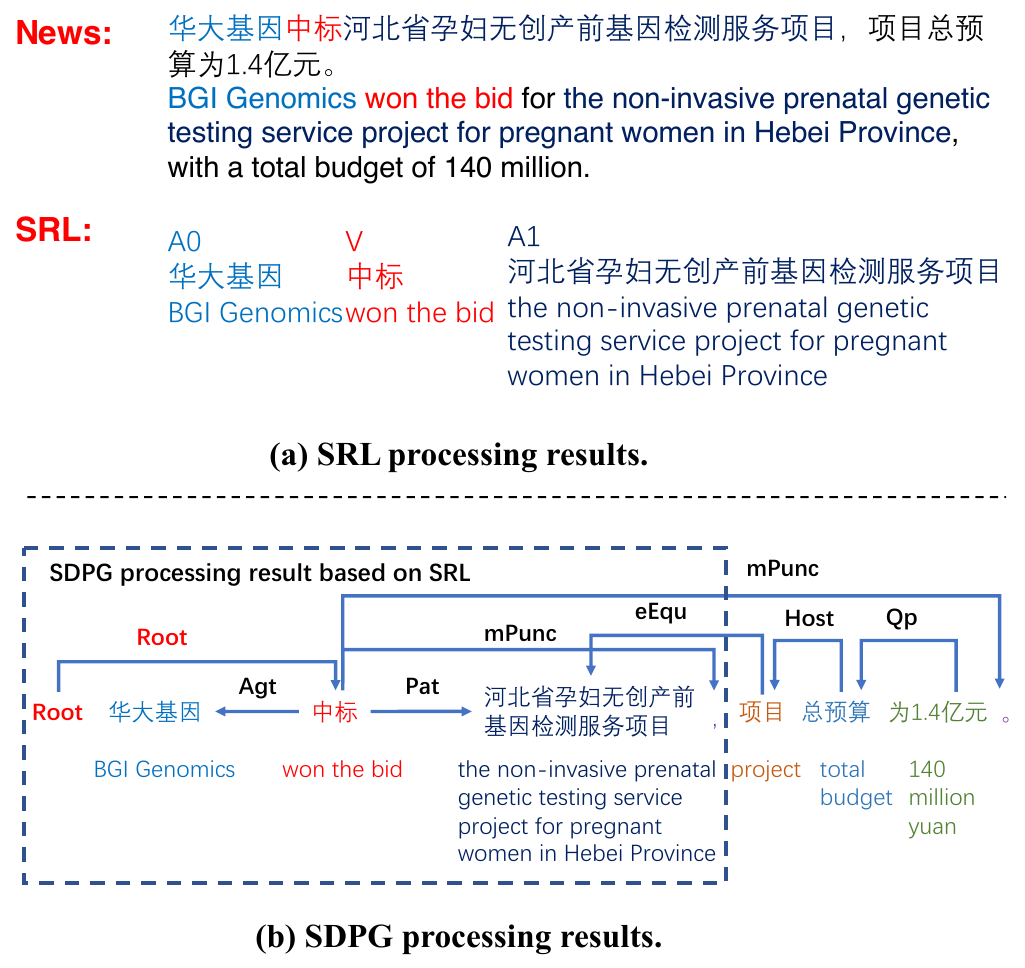}
    \vspace{-0.3cm}
    \caption{The processing results of SRL and SDPG.
    }
    \label{fig:exactionNews}
    \vspace{-0.5cm}
\end{figure}

In terms of financial news, we consider both the overall semantic information and the roles information within the news sentence. Specifically, we obtain the semantic information via a pre-trained textual encoder (i.e. RoBERTa), while the roles information is extracted through SRL and SDPG. SRL annotates the semantic roles,  the verb (V), proto-agent (A0), and proto-patient (A1) as word embeddings, within the financial news, where $e_{V}^{i}$ represents the token index of $V$ in sentence $i$. Further, we utilize SDPG  to construct a semantic dependency parsing graph. As shown in Fig. \ref{fig:exactionNews}, every relationship was showed as their grammar categorical names. For simplicity, it is denoted as $\mathbf{X}_{A B}$, which represent the edge feature of connecting two roles in semantic dependency graph. Subsequently, we employ a pooling operation to aggregate the semantic roles and dependency graph. For instance, the pooling V1 factor can be formulated as follows:
\begin{equation}
    \begin{aligned}
        \mathbf{e}^{SRL}  & =\operatorname{pooling}\left(\left\{\mathbf{e}^{SRL}\right\}\right)  \\
        \mathbf{e}^{SDPG} & =\operatorname{pooling}\left(\left\{\mathbf{e}^{SDPG}\right\}\right) \\
    \end{aligned},
\end{equation}
where $\mathbf{e}^{SRL}$ and $\mathbf{e}^{SDPG}$ are the final representation of the financial news. Denoting the edge attribute in semantic graphs between the semantic roles as $\mathbf{G}_{V A0}$, $\mathbf{G}_{V A1}$, $\mathbf{G}_{A0 A1}$. Based on this, an input feature matrix $\mathbf{X}$ contain $N$ pieces of news can be formulated as follows:
\begin{equation}
    \mathbf{X}_{n}=\left[\begin{array}{ccc}
            \mathbf{e}_{v}^{1}    & \dots & \mathbf{e}_{v}^{N}    \\
            \mathbf{e}_{A  0}^{1} & \dots & \mathbf{e}_{A  0}^{N} \\
            \mathbf{e}_{A  1}^{1} & \dots & \mathbf{e}_{A  1}^{N} \\
            \mathbf{G}_{V A0}^{1} & \dots & \mathbf{G}_{V A0}^{N} \\
            \mathbf{G}_{V A1}^{1} & \dots & \mathbf{G}_{V A1}^{N} \\
            \mathbf{G}_{A0A1}^{1} & \dots & \mathbf{G}_{A0A1}^{N} \\
        \end{array}\right].
\end{equation}

For stock factors, which represents the intrinsic characteristics of the stock such as size, value and momentum, it is represented as a matrix $\mathbf{X}_{F}$ as follows, where $f_m$ is the $m$-th factor of the stock.
\begin{equation}
    \mathbf{X}_{f}=\left[\begin{array}{ccc}
            f_{1}^{1} & \dots  & f_{1}^{N} \\
            f_{2}^{1} & \dots  & f_{2}^{N} \\
            \vdots    & \ddots & \vdots    \\
            f_{m}^{1} & \dots  & f_{m}^{N} \\
        \end{array}\right].
\end{equation}

To this end, we concatenate the news and stock factors to obtain the complete factor matrix $\mathbf{X}$ as follows:
\begin{equation}
    \mathbf{X}=\mathbf{W}_{\alpha} \odot \left[\begin{array}{c}
            \mathbf{X}_{n} \\
            \mathbf{X}_{f}
        \end{array}\right],
\end{equation}
where $W_{\alpha}$ is an adaptive weight to balance the importance of news and stock factors. With aforementioned features, we use a 2-layers MLP and a softmax layer to predict the classification of returns as follows:
\begin{equation}\label{eqn:mlp}
    \mathbf{y}_n= Softmax(MLP(\mathbf{X}_{[:,n]})).
\end{equation}


The definition of classification $\mathbf{y}_n$ in the ~\citet{zou-etal-2022-astock} is followed, where mainly divided into three categories: positive, netural, and negative:
\begin{equation}\label{return}
    \begin{aligned}
        \operatorname{E}=\left\{\begin{array}{ll}
                                          \text { positive } & \text { if } \operatorname{r} \text{ ranked top } 20 \%          \\
                                          \text { neutral }  & \text { if } \operatorname{r}  \text{ ranked }  20 \% \sim 60 \% \\
                                          \text { negative } & \text { if } \operatorname{r} \text{ ranked bottom }  20 \%
                                      \end{array}\right.
    \end{aligned},
\end{equation}
where $r$ is the ranking of one stock's return ratio in the market. This module is trained from cross entropy loss, which can be format as:
\begin{equation}
    \mathcal{L}_{nf} = CrossEntropy(\mathbf{y},\mathbf{E}),
\end{equation}

\subsection{Return Forecasting Module}
\label{subsec:ff5n}
In order to comprehensively analyze and predict the impact of news on stock prices, we have introduced news factors based on the Fama-French 5 factors model\cite{FAMA20151}, leading to the creation of the FF5-News model. Within this model, we utilize the aforementioned News Factorization Module to obtain news classifications represented as $\mathbf{y}$, and incorporate specific stock trading data as inputs to the model. This process results in the establishment of FF5-News factors, enabling a more comprehensive multidimensional forecast of stock returns.

The Fama-French 5 factors model offers multiple dimensions for predicting and analyzing the intricacies of the stock market. It is widely utilized in academia and the finance industry for research and investment decision-making purposes.The model is expressed as follows:
\begin{equation}
    \begin{aligned}
        R_{i t}-R_{f t} & =                                                                     \alpha_{i}+\beta_{i}\left(R_{m t}-R_{f t}\right)+b_{i}SMB_{t}                                                        \\
                        & +h_{i}HML_{t}+r_{i}RMW_{t}                                                                                                                                 +c_{i}CMA_{t}+\varepsilon_{i t}
    \end{aligned}
\end{equation}
where $R_{it}$ is the extra return of stock $i$ at time $t$, $R_{ft}$ is the risk-free rate at time $t$, $R_{mt}$ is the market return at time $t$, and $\alpha_{i}$ is the intercept of the regression model. $\beta_{i}$ is the market factor that measures the sensitivity of stock $i$ to the market factor. $SMB_{t}$ is the size factor, $HML_{t}$ is the valuation factor, $RMW_{t}$ is the profit factor, $CMA_{t}$ is the investment factor. $b_{i}$, $h_{i}$, $r_{i}$, $c_{i}$, $u_{i}$ are the exposure coefficients of the corresponding factors, and $\varepsilon_{it}$ is the zero-mean random term which represents the residual randomness of the stock. The target of this model is to minimize the $\alpha_{i}$.

However, the Fama-French 5 factors model can't sensitively capture the changes in stock returns caused by news announcements. Therefore, we introduce FF5-News, an integrated model that enhances precision and explanatory power by incorporating news factors.The model is expressed as follows:
\begin{equation}
    \begin{aligned}
        R_{i t}-R_{f t} & =                                                                     \alpha_{i}+\beta_{i}\left(R_{m t}-R_{f t}\right)+b_{i}SMB_{t}                                      \\
                        & +h_{i}HML_{t}+r_{i}RMW_{t}                                                                                                                                 +c_{i}CMA_{t} \\
                        & +m_{i}M_{t}+\varepsilon_{i t}
    \end{aligned}.
\end{equation}
where $M_{t}$ is the news factor, $m_{i}$ is the exposure coefficients of the news factor.

According to the factor definitions in the Fama-French 5 factors model, we believe that news articles categorized $\mathbf{y}$ as 'Positive' within the News Factorization Module will yield higher returns compared to those categorized $\mathbf{y}$ as negative.Therefore, we adopted the approach recommended by Fama and French in their paper\cite{FAMA20151}, using the news classification $\mathbf{y}$ obtained from the News Factorization Module as a profitability indicator in the news dimension. This successfully quantifies the impact of news on stock returns, resulting in a metric named $M_{t}$. By integrating $M_{t}$ with the Fama-French 5 factors model, the FF5-News model was established. For specific construction methods, please refer to the Appendix:B.
Finally, through our FF5-News model, we can get a comprehensive return yield prediction in terms of market risk, size, valuation, profitability, investment style, and news effect, which has stronger explainable.



\subsection{Risk Assessment Module}
To provide a comprehensive FinReport, it is essential to evaluate and manage risks in addition to the returns dimension described above. To achieve this, we develop a VaR risk assessment system that supplements our model with risk analysis dimension.To calculate the VaR risk, we follow \cite{VillarRubio2023} 
 and construct an EGARCHmodel to fit the time-varying volatility of the stock, which captures the volatility of financial time series, especially the heteroscedasticity of volatility, i.e., volatility clustering in the financial market. The conditional equation of the EGARCH (p, q) model is as follows:
\begin{equation}
    \begin{aligned}
        \ln \sigma_{t}^{2} & =\omega+\sum_{l=1}^{p} \alpha_{l}\left(\left|e_{t-l}\right|-\mathbb{E}\left|e_{t-l}\right|\right) \\
                           & +\sum_{j=1}^{q} \beta_{j} \ln \sigma_{t-j}^{2} + \sum_{k=1}^{q} \gamma_{k} e_{t-k}^{2}
    \end{aligned},
\end{equation}
where $\sigma_{t}^{2}$ is the volatility (variance) at time $t$, $e_{t}$ is the residual term at time $t$, $\omega$ is the constant term of the model, $\alpha_{l}$ is the weight of the residual. And $\beta_{j}$ is used to consider the autoregressive characteristics of the volatility, $\gamma_{k}$ is used to consider the ARCH effect of the volatility (the impact of volatility on itself).

Subsequently, the volatility is derived via the fitted EGARCH model, and the VaR estimate is computed by amalgamating the Z quantile of the confidence level (provided by the standard normal distribution). The VaR estimate represents the highest potential loss within a future horizon, and is formulated as follows:
\begin{equation}
    \operatorname{VaR}= \mu_{t}-\sigma_{t} \times Z_{\alpha},
\end{equation}
where $\mu_{t}$ is the expected return at time $t$, $\sigma_{t}$ is the volatility of the stock at time $t$, and $Z_{\alpha}$ is the Z quantile of the confidence level $\alpha$.

The loss of VaR module can be calculated by the following equation:
\begin{equation}
    \mathcal{L}_{var}=\frac{1}{N} \sum_{i=1}^{N} | \operatorname{VaR}_{i}-\operatorname{ActualVaR}_{i} |,
\end{equation}
where $N$ is the number of stocks, $\operatorname{VaR}_{i}$ is the predict VaR value of the $i$-th stock, and $\operatorname{ActualVaR}_{i}$ is the actual VaR value of the $i$-th stock.

With aforementioned VaR assessment, the estimated interval of the maximum loss of the stock is obtained to help our report system control the risk.

\subsection*{Report Generation Based On LLM}

\begin{table*}[t]
\centering
\caption{Performance comparison of different models}
\vspace{-0.2cm}
\begin{tabular}{lllcccc}
\hline
\hline
\multirow{2}{*}{Model} &
  \multirow{2}{*}{Reference} &
  \multirow{2}{*}{Resource} &
  \multirow{2}{*}{Accuray} &
  \multirow{2}{*}{F1 Score} &
  \multirow{2}{*}{Recall} &
  \multirow{2}{*}{Precision} \\
                                    &                                      &              &       &       &       &       \\ \hline
StockNet \cite{Xu2018}                            & ACL 2018                        & News         & 46.72 & 44.44 & 46.68 & 47.65 \\
HAN Stock \cite{Hu2018}                           & WSDM 2018                        & News         & 57.35 & 56.61 & 57.20 & 58.41 \\
Bert Chinese \cite{Devlin2019}                       & ACL 2019                   & News         & 59.11 & 58.99 & 59.20 & 59.07 \\
ERNIE-SKEP \cite{Tian2020}                          & ACL 2020                     & News         & 60.66 & 60.66 & 60.59 & 61.85 \\
XLNET Chinese \cite{Cui2020}                        & EMNLP 2020                     & News         & 61.14 & 61.19 & 61.09 & 61.60 \\
\multirow{2}{*}{RoBERTa WWM Ext \cite{Cui2020}  }    & \multirow{2}{*}{EMNLP 2020  }            & News         & 61.34 & 61.48 & 61.32 & 61.97 \\
                                    &                                      & News+Factors & 62.49 & 62.54 & 62.51 & 62.59 \\
\multirow{2}{*}{Chinese Lert Large \cite{cui2022lert}} & \multirow{2}{*}{Arxiv 2022} & News         & 64.37 & 64.30 & 64.31 & 64.34 \\
                                    &                                      & News+Factors & 66.36 & 66.16 & 66.69 & 66.4  \\
\multirow{2}{*}{Chinese Pert Large \cite{cui2022pert}} & \multirow{2}{*}{Arxiv 2022} & News         & 65.09 & 65.03 & 65.07 & 65.02 \\
                                    &                                      & News+Factors & 67.37 & 67.27 & 67.73 & 67.28 \\
\multirow{3}{*}{Self-supervised SRLP \cite{zou-etal-2022-astock}} &
  \multirow{3}{*}{FinNLP 2022} &
  Factors &
  59.76 &
  59.71 &
  59.75 &
  59.72 \\
                                    &                                      & News         & 62.97 & 63.05 & 62.93 & 63.47 \\
                                    &                                      & News+Factors & 66.89 & 66.92 & 66.95 & 66.92 \\ \hline
Factors only                        & \multirow{5}{*}{Ours}                & Factors      & 63.74 & 63.66 & 63.71 & 63.67 \\
SRL\&SDPG                           &                                      & News         & 66.10 & 66.01 & 66.09 & 66.04 \\
SRL \& Factors                      &                                      & News+Factors & 69.48 & 69.28 & 69.41 & 69.54 \\
SDPG \& Factors                     &                                      & News+Factors & 73.12 & 72.97 & 72.96 & 73.04 \\
SRL \& SDPG \& Factors              &                                      & News+Factors & 75.40 & 75.12 & 75.23 & 75.42 \\ \hline \hline
\end{tabular}
    \vspace{-0.2cm}
    \label{tab:main}
    
\end{table*}

In the aforementioned module, we have obtained multidimensional stock return prediction analysis data and risk assessment data. To this end, we propose a report generation model based on a large language model (LLM). Consequently, the final report can be generated, with the following format:
\begin{equation}
    \text{report} = LLM(R,M, \operatorname{VaR})
\end{equation}
where R is aforementioned multidimensional stock Earning. We use prompt M for LLM which is \textit{``Based on multi-dimensional predictive information and risk assessment values, a financial analysis report will be generated, comprising four main sections: return forecasting, risk assessment, overall trend prediction, and summary. Among them, the return forecasting section is required to include predictive analyses in six dimensions: Market Factor, Size Factor, Valuation (BP) Factor, Profitability Factor, Investment Factor, and News Effect Factor. The risk assessment section provides an estimation of the maximum potential loss, while the overall trend prediction outputs either `Positive' or `Negative' based on the overall profitability. The summary section includes a comprehensive analysis of the predictive information and risk assessment, offering an integrated evaluation of the investment potential of the stock.''}


\section{Experiments}\label{sec:exper}


\subsection{Datasets}
\textbf{Astock}\footnote{\url{https://github.com/JinanZou/Astock}} is a dataset proposed by \cite{zou-etal-2022-astock}, which contains stock and news data from 2018-07-01 to 2021-11-30. Specifically, this famous dataset spanning 1248 days contains many items, such as price, news, and so on. It's the only dataset which contains both factors and news. We follow the splitting strategy of \cite{zou-etal-2022-astock}.  2018-07-01 to 2020-7-01 of the dataset is used for training, 2020-7-01 to 2020-10-01 for validation, and the rest for testing (2020-10-01 to 2020-12-31).

\subsection{Evaluation Metrics}
Firstly, we utilized accuracy, F1 score, recall, and precision as evaluation metrics for our news factorization module, same as \cite{zou-etal-2022-astock}. Secondly, we employed the GRS (Goodness-of-Fit R-squared)\cite{COLINCAMERON1997329} metric, its p-value and the mean absolute value of alpha to measure the model's explanatory power on return predictions for explainable news effects. For VaR (Value at Risk) risk assessment, we used RMSE (Root Mean Squared Error), MAE (Mean Absolute Error), and VaR Loss Coverage Rate as evaluation metrics. The evaluation metrics for the report generation system are currently omitted. To simulate trading based on the generated analysis reports, we utilized annualized rate of return, maximum drawdown, and Sharpe ratio as evaluation metrics.

\subsection{Implementation Details}
For specific hyper-parameters in our method, all baselines are compared with same hyperparameters search grid, [$1e^{-3}$,$1e^{-4}$,$1e^{-5}$,$1e^{-6}$] of learning rate, [16,32,64,128] of batch size. The dropout rate is set to be 0.1, while the hidden size of our two layers MLP in Eqn.\ref{eqn:mlp} is $1,024$.

\subsection{News Factorization Evaluation}
To evaluate the effectiveness of news factorization, we compared our model with two kinds of state-of-the-art baselines. 1) large pre-trained language models, including StockNet \cite{Xu2018}, HAN Stock \cite{Hu2018}, Bert Chinese \cite{Devlin2019}, ERNIE-SKEP \cite{Tian2020}, and XLNET Chinese \cite{Cui2020}. 2) methods are integrated both news and factors information, including RoBERTa WWM Ext \cite{Cui2020}, Chinese Lert large \cite{cui2022lert}, Chinese Pert large \cite{cui2022pert}, and Self-supervised SRLP \cite{zou-etal-2022-astock}. 

The experimental results are presented in Table \ref{tab:main}. The following observations can be made: 1) The combination of factor and news approaches, generally outperform factor-only (e.g. Self-supervised SRLP) and news-only models (e.g. Chinese Pert Large). 2) SRLP using a more powerful embedding model achieves excellent performance, probably because it obtains a reasonable representation of the news. 3) Our model outperforms all the other methods, manifests that our model focus on better extracting the news information with semantic roles and relationship awareness and combining it with the factor data. For the above Pert pre-trained language models, we extracted embeddings from the [CLS] token and attached a three-classifier to predict stock trends.

To better understand the contributions of different components in our framework, we conduct component studies on multiple variants: 1) News; 2) Factors; 3) SRL; 4) SDPG.
Specifically, we explored four scenarios within our model framework: extracting factors information using only the MLP method, extracting news information using only the SRL method, using only the SDPG method, and using a combination of both SDPG and SRL. The detailed evaluation results are shown in Table ~\ref{tab:main}. From the results, we draw the following two observations: 1) Our method after combining factors and news significantly outperforms factor-only and news-only variants. This fact may arise from factor information is not able to sensitively capture the fluctuations in stock returns caused by changes in news, and relying solely on news information without considering the inherent characteristics of the stocks also cannot fully explain the variations in stock returns. 2) After ablation of the SRL and SDPG components, the model performance degrades, suggesting that every piece of our model is essential. Among them, the ablation SRL represents that our model cannot extract a concise representation of key information from the news, while  the ablation SRL represents that our model cannot capture the semantic dependencies among essential information within the news.

\begin{table}[t]
  \centering
  \caption{Fama-French 5 Factors with News Effect Factor model GRS test results and VaR risk assessment system evaluation results}
  \begin{adjustbox}{width=\columnwidth,center}
    \begin{tabular}{llll}
      \hline
      \hline
      Model                                                                                    &
      GRS                                                                                      &
      GRS p-value                                                                              &
      \begin{tabular}[c]{@{}l@{}}Mean Absolute\\ Value of Alpha\end{tabular}                                                              \\ \hline
      Fama-French 5 Factors                                                                    &
      3.8345                                                                                   &
      $1.791\times 10^{-3}$                                                                    &
      0.0754                                                                                                                              \\
      \begin{tabular}[c]{@{}l@{}}Fama-French 5 Factors \\ with News Effect Factor\end{tabular} & 5.3526 & $0.000\times 10^{-3} $ & 0.0602 \\ \hline \hline
      Model                                                                                    &
      RMSE                                                                                     &
      MAE                                                                                      &
      \begin{tabular}[c]{@{}l@{}}VaR Loss \\ Coverage Rate\end{tabular}                                                                   \\ \hline
      \begin{tabular}[c]{@{}l@{}}VaR Risk Assessment\\  System\end{tabular}                    &
      0.0947                                                                                   &
      0.8176                                                                                    &
      0.8123                                                                                                                              \\ \hline
      \hline
    \end{tabular}
  \end{adjustbox}
  \label{tab:fffive}
    \vspace{-0.85cm}
\end{table}

\subsection{Return Forecasting Research}
In order to assess the effectiveness of the newly proposed News Effect Factor in this paper with precision and to validate the model's explanatory capabilities after the incorporation of this News Effect Factor in real-world, we have opted to out of distribution (OOD) data for in-depth analysis and validation of the model.Specifically, we selected all the data with stock news announcements in the Astock financial dataset from 2021-01-01 to 2021-11-01. 
The GRS indicators of FF5 and FF5-News models are compared in Table ~\ref{tab:fffive}.

From Table ~\ref{tab:fffive}, it is evident that the FF5-News model has a better GRS indicator, with a lower p-value, mean absolute value of alpha, and a higher GRS value. These results demonstrate that our model has a strong explanatory power for predicting returns and outperforms the Fama-French 5 Factors model. Therefore, the News Effect Factor we constructed can effectively explain a portion of the excess return rate and supplement the unexplained part of the FF5 model, thereby enhancing the explanatory power and predictive analysis ability.

\begin{table}[t]
    \caption{Backtest in real-world scenarios}
  \begin{adjustbox}{width=\columnwidth,center}
        \begin{tabular}{llll}
        \hline
        \hline
            \multirow{2}{*}{Model}                                                               &
            \multirow{2}{*}{\begin{tabular}[c]{@{}l@{}}Maximum\\ Drawdown\end{tabular}}          &
            \multirow{2}{*}{\begin{tabular}[c]{@{}l@{}}Annualized Rate\\ of Return\end{tabular}} &
            \multirow{2}{*}{\begin{tabular}[c]{@{}l@{}}Sharpe\\ Ratio\end{tabular}}                                              \\ 
                                                                                                 &          &          &         \\ \hline
            CSI300                                                                                 & -18.19\% & -9.65\% & -0.3653 \\ 
            XIN9                                                                                 & -26.70\% & -18.36\% & -0.6253 \\ 
            \begin{tabular}[c]{@{}l@{}}Chinese-PERT-large + \\ Factors\end{tabular}              & -4.17\%  & 11.26\%  & 1.9621  \\ 
            \begin{tabular}[c]{@{}l@{}}Self-supervised SRLP + \\ Factors\end{tabular}            & -3.66\%  & 26.38\%  & 3.76    \\ 
            \begin{tabular}[c]{@{}l@{}}SRL \& SDPG + \\ Factors\end{tabular}                     & -3.06\%  & 57.76\%  & 7.4043  \\ 
                    \hline
                    \hline
        \end{tabular}
  \end{adjustbox}
    \label{tab:backtest}
    \vspace{-0.6cm}
\end{table}

\subsection{VaR Risk Assessment}
In this section, we assess the effectiveness of our VaR risk assessment module. The results are presented in Table ~\ref{tab:fffive}. The results indicate that the RMSE indicator of our risk assessment system is 0.0947, and the MAE is 0.817. These two indicators demonstrate that our risk system has a high level of predictability for risk assessment and can provide a valuable source of information for risk analysis. Furthermore, our VaR loss coverage rate indicator achieved an impressive 0.8123, which confirms that our system covers a significant portion of the losses and has a strong risk reference significance.

\begin{figure*}[t]
    \centering
    
    \includegraphics[width=0.9\textwidth]{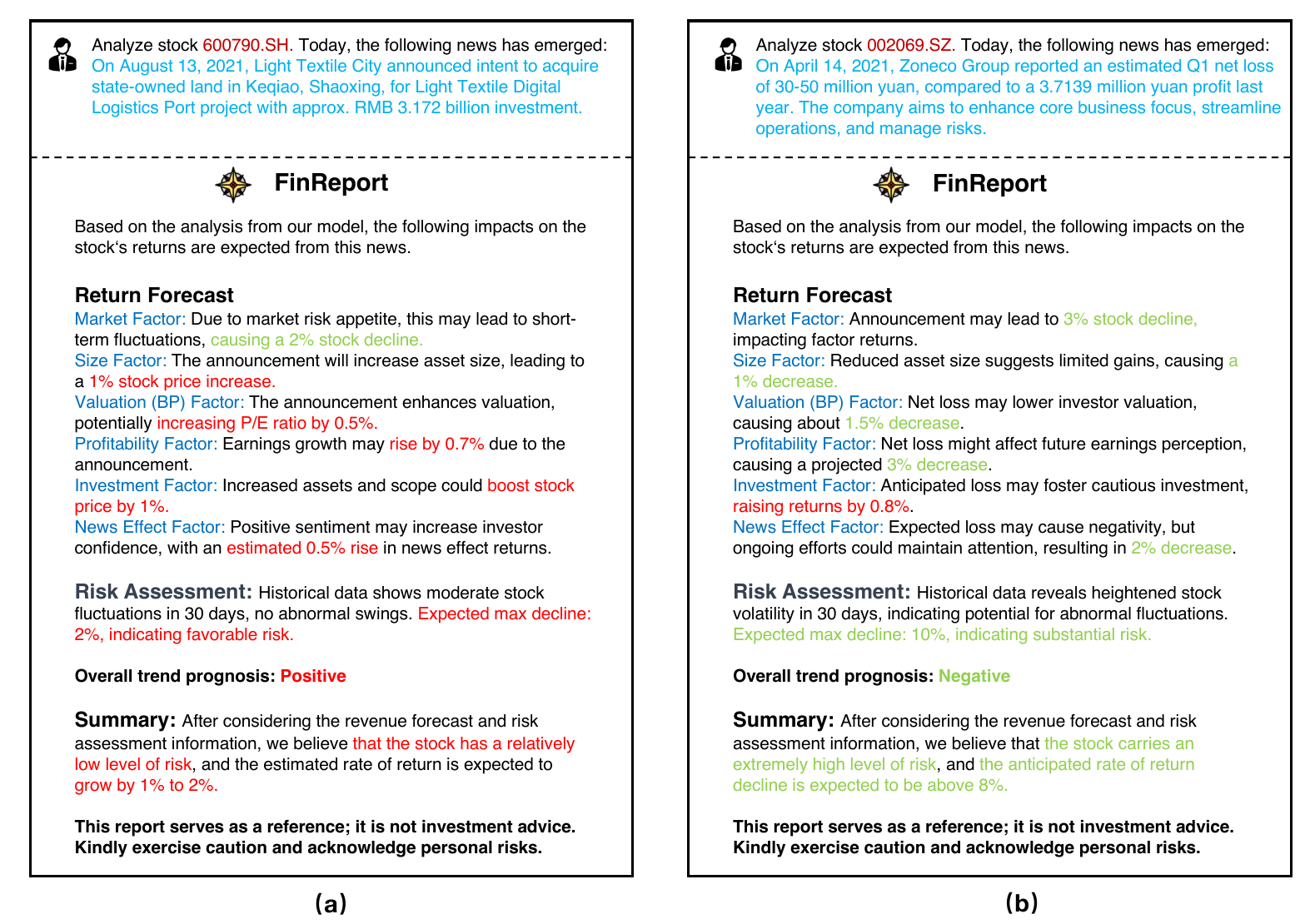}
    \vspace{-0.2cm}
    \caption{Examples of both postive and negateive report}
    \vspace{-0.3cm}
    \label{fig:report}
    
\end{figure*}

\subsection{Report Generation}
In this section, we utilized the LLMs (e.g. ChatGPT) to generate comprehensive reports for users, in which the multi-dimensional return forecasting results and VaR risk assessment results will be used as input. Figure ~\ref{fig:report} illustrates two specific examples of our FinReport. 

As shown in Figure ~\ref{fig:report}(a), for a good news announcement (i.e. \textit{Light Textile City announced intent to acquire state-owned land in Keqiao}), the investment advice given by our FinReport is ``positive''. Among them, the return forecasting is composed of multiple dimensions such as stock factors and size factors, while the risk assessment believes that it is only 2\%. Conversely, for bad news as shown in Figure ~\ref{fig:report}(b), the return forecasting is negative and the risk assessment reaches 10\%.

Further, to evaluate users' satisfaction with our generated reports, we invited 10 experts to rate the $1,000$ generated reports from three levels, namely -1, 0, 1. In the end, due to the factuality and high fluency of FinReport, users gave 96\% positive score, which shows that the automatically generated reports are more convenient than self-finding information. Of course, the premise is that it is sufficiently convincing, which is what this paper is dedicated to.

\subsection{Profitability Evaluation In Real-world scenarios}
\begin{figure}[t]
    \centering
    \includegraphics[width=0.48\textwidth]{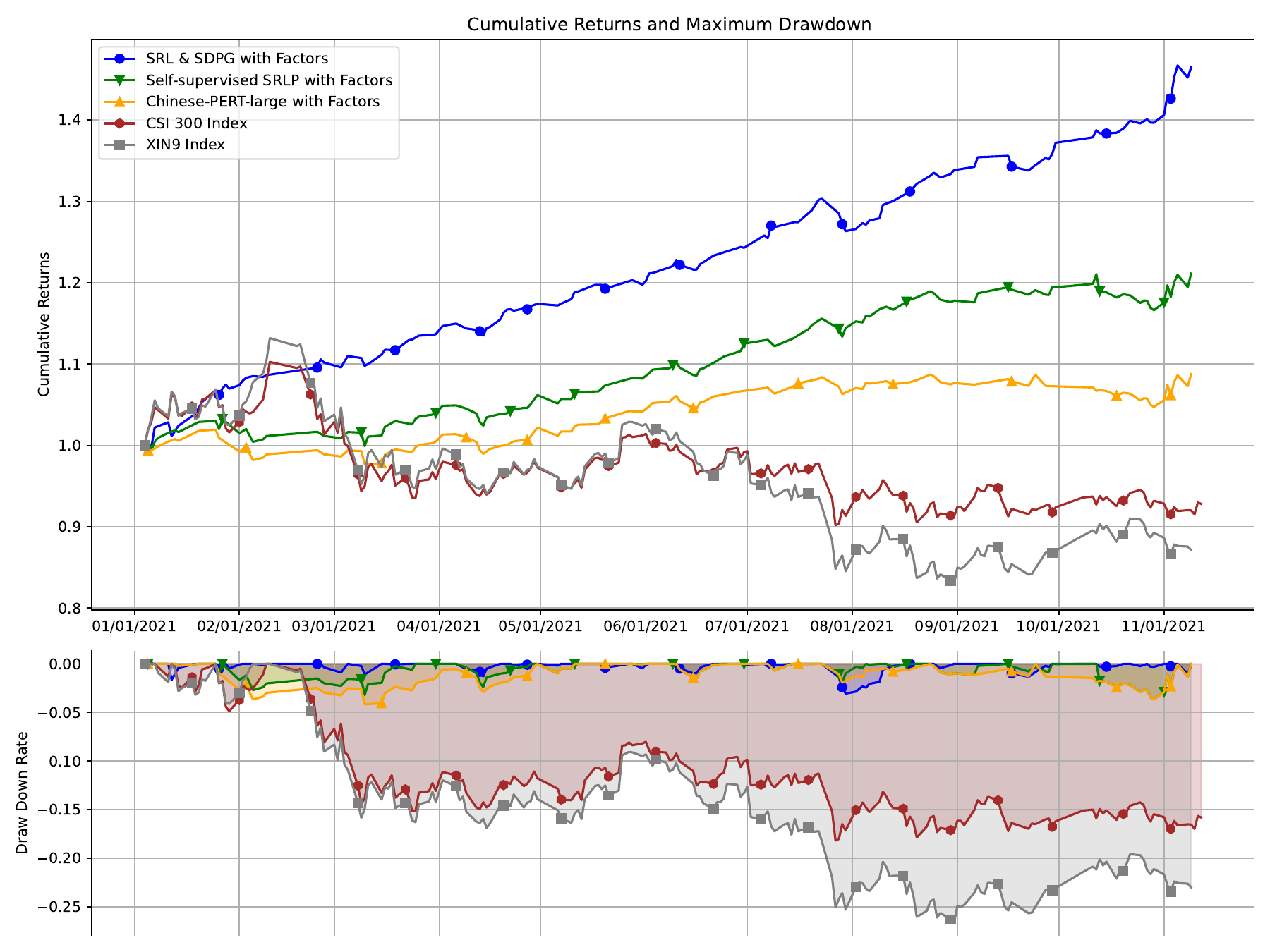}
    \caption{Comparing the performance of different models in real-world scenarios}
    \label{fig:backtest}
    \vspace{-0.5cm}
\end{figure}

In this section, we evaluate our method in real-world by backtest. In the view of investment, we hope that our model can achieve on out of distribution (OOD) generalization. To achieve this goal, we conducted simulated trading from 2021-01-1 to 2021-11-01.

Notably, considering the trading restrictions in China's A-share market, we bought the constructed portfolio at the opening price on each trading day and sold it at the opening price on the next trading day, assuming a transaction cost of 0.1\% for each trade.

%

In Table ~\ref{tab:backtest}, we can observe that our approach achieved a remarkable annualized rate of return of 57.76\%, surpassing previous baselines and market indices XIN9 and CSI300. 
Additionally, our method obtained the lowest Maximum Drawdown of -3.06\% and the highest Sharpe Ratio of 7.4043, significantly outperforming previous methods and indicating that our self-supervised approach successfully achieved higher expected returns while maintaining relatively lower risk, as shown in Figure ~\ref{fig:backtest}.

\section{Conclusions}\label{sec:conclus}
In this paper, we present FinReport, a novel model for generating financial reports. Our FinReport is designed to ensure accuracy, professionalism, and comprehensiveness, and consists of three sub-modules: News Factorization, Return Forecasting, and Risk Assessment. Leveraging innovative techniques, we utilize Propbank-style semantic role labeling (SRL) and semantic dependency parsing graphs (SDPG) to extract key information from news articles and combine it with stock features. The incorporation of news factors into the Fama-French 5-factor model improves stock return prediction. Our experimental results demonstrate superior performance compared to benchmarks, providing investors with more insightful and comprehensive financial news analysis reports.

\section{Acknowledgments}
This work is supported in part by the National Natural Science Foundation of China (62372187), in part by the National Key Research and Development Program of China (2021YFC2202603) and in part by the Guangdong Provincial Key Laboratory of Human Digital Twin (2022B1212010004).

\newpage
\bibliographystyle{ACM-Reference-Format}
\bibliography{sample-authordraft}

\newpage
\section{appendix}
In this appendix, to further analyze the effectiveness of our proposed method, more details about the experiment settings and results are presented.

\subsection{Return Calculation}
\label{App:A}

Groups in FF5 Model are written as follows:

\begin{itemize}[leftmargin=*]
\item \textbf{Size}: big/small
\item \textbf{BP}: High/Netural/Low
\item \textbf{Profitability}: Robust/Neutral/Weak
\item \textbf{Investment}: Conservative/Neutral/Aggressive
\end{itemize}
Groups in FF5-News Model are written as follows:

\begin{itemize}[leftmargin=*]
\item \textbf{Size}: big/small
\item \textbf{BP}: High/Netural/Low
\item \textbf{Profitability}: Robust/Neutral/Weak
\item \textbf{Investment}: Conservative/Neutral/Aggressive
\item \textbf{NewsEffect}: Positive/Medium/Negative
\end{itemize}

\subsection{Factor Construction}
\label{App_B}
Factors in FF5 Model are formulated as follows:
\begin{equation*}
    \begin{split}
        \begin{aligned}
            SMB_{BP}  & = (SH+SN+SL)/3 - (BH+BN+BL)/3,      \\
            SMB_{op}  & =(SR+SN+SW)/3 -(BR+BN+BW)/3,        \\
            SMB_{inv} & =(SC+SN+SA)/3 -(BC+BN+BA)/3,        \\
            SMB       & =(SMB_{BP}+ SMB_{op}+ SMB_{inv})/3, \\
            HML       & = (BH+SH)/2 - (BL+SL)/2,            \\
            RMW       & = (BR+SR)/2-(BW+SW)/2,             \\
            CMA       & = (BC+SC)/2 – (BA+SA)/2.            \\
        \end{aligned}
    \end{split}
\end{equation*}

Factors in FF5-News Model are formulated as follows:
\begin{equation*}
    \begin{split}
        \begin{aligned}
            SMB_{BP}   & =(SH+SN+SL)/3 -(BH+BN+BL)/3,      \\
            SMB_{op}   & =(SR+SN+SW)/3 -(BR+BN+BW)/3,      \\
            SMB_{inv}  & =(SC+SN+SA)/3 -(BC+BN+BA)/3,      \\
            SMB_{news} & =(SP+SM+SN)/3 -(BP+BM+BN)/3,      \\
            SMB        & = (SMB_{BP}+ SMB_{op}+ SMB_{inv}
            + SMB_{news})/4,                               \\
            HML        & = (BH+SH)/2 - (BL+SL)/2,          \\
            RMW        & = (BR+SR)/2-(BW+SW)/2,            \\
            CMA        & = (BC+SC)/2 – (BA+SA)/2,          \\
            News       & = (BP+SP)/2 – (BN+SN)/2.          \\
        \end{aligned}
    \end{split}
\end{equation*}

\subsection{Detial infomation of back test}
In this section, we provide the more widely used metrics and the details of backtest in Figure 1,2,3. Important metrics are described as below.

\newpage
\begin{figure}[]
    \centering
    \includegraphics[width=0.9\linewidth]{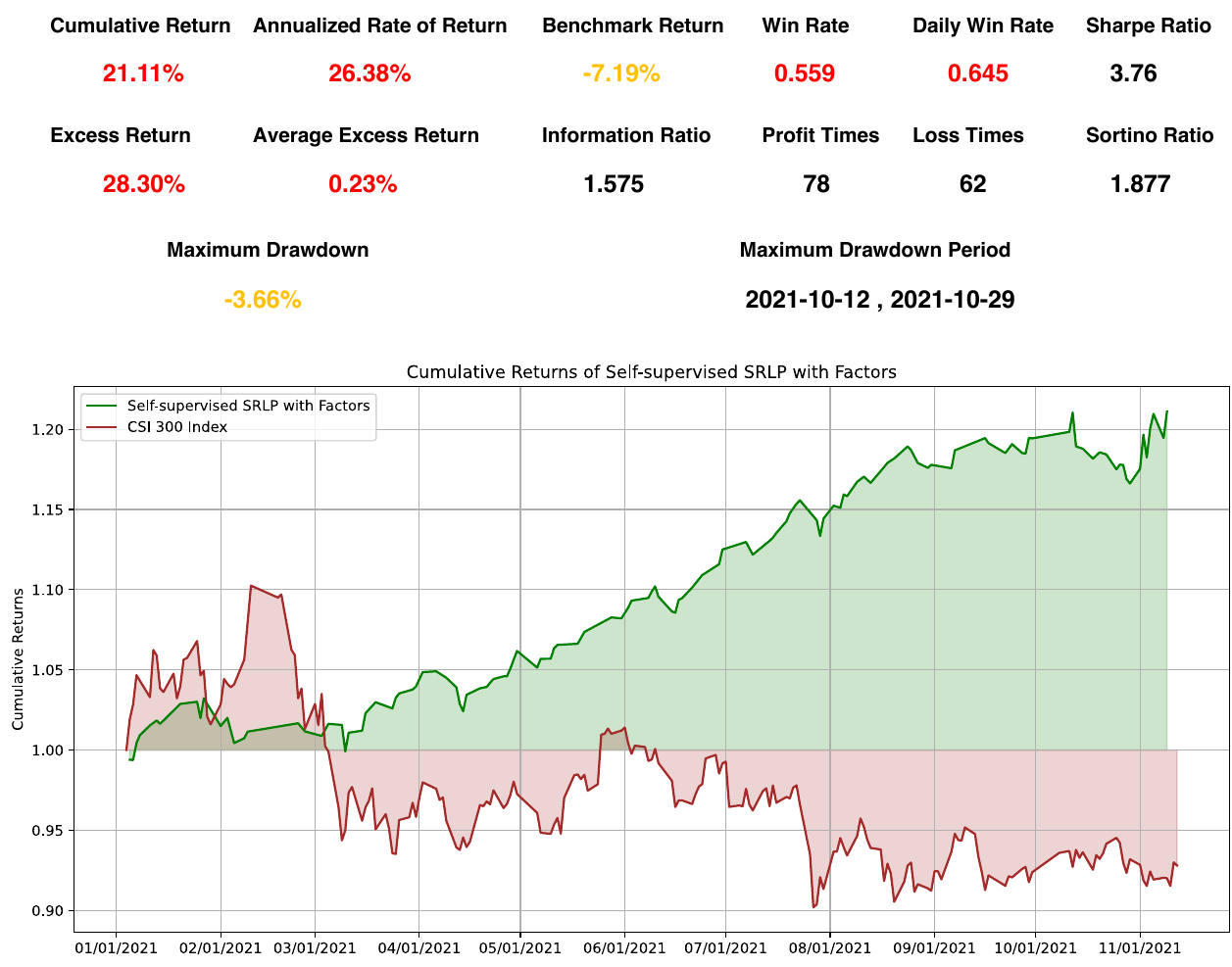}
    \caption{Details of backtest on Self-supervised SRLP with factors}
    \vspace{-0.5cm}
    \label{fig:appendix2}
\end{figure}
\begin{figure}[]
    \centering
    \includegraphics[width=0.9\linewidth]{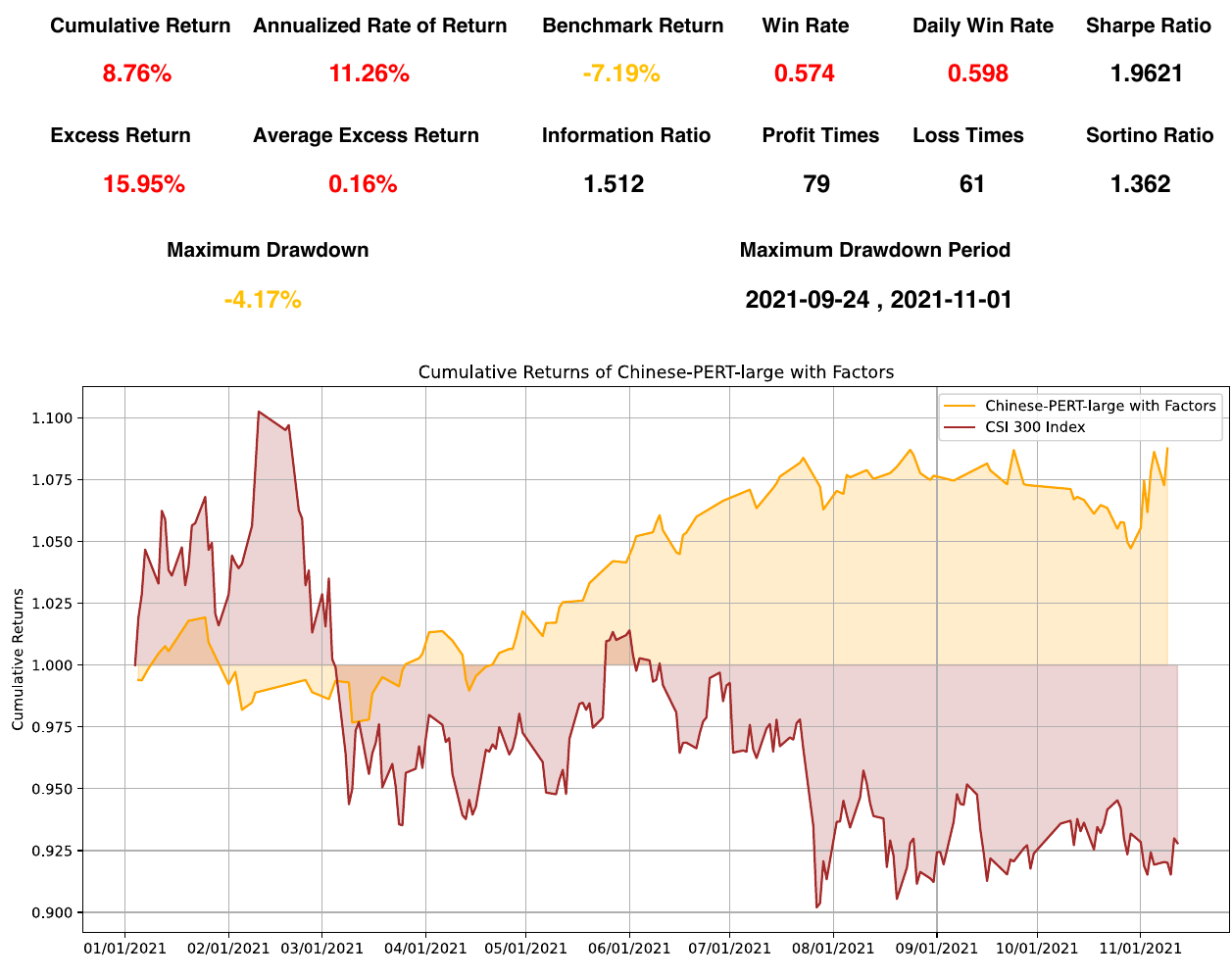}
    \caption{Details of backtest on Chinese Pert large}
    \vspace{-0.5cm}
    \label{fig:appendix3}
\end{figure}
\begin{figure}[]
    \centering
    \includegraphics[width=0.9\linewidth]{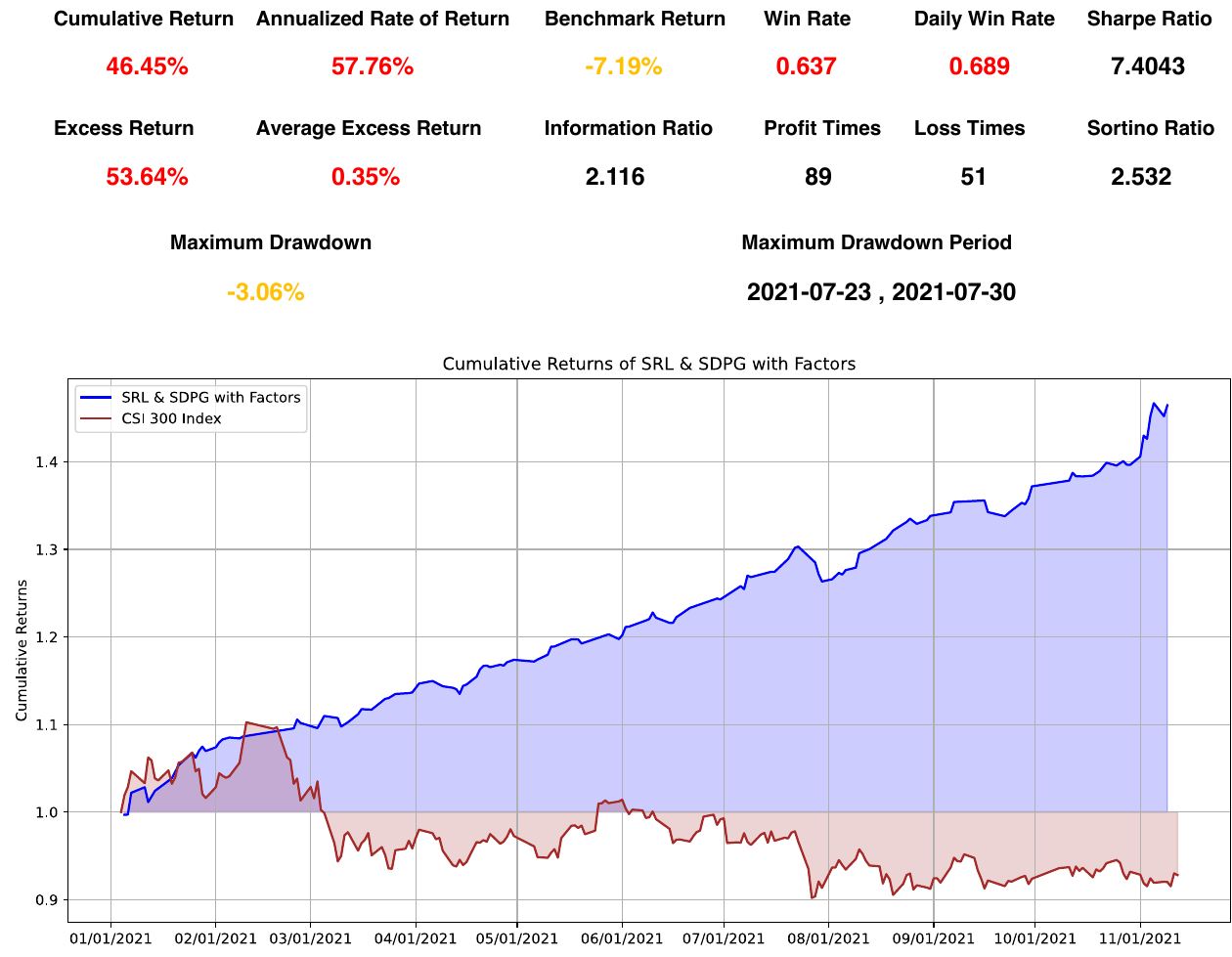}
    \caption{Details of backtest on SRL\& SDPG with factors}
    \vspace{-0.5cm}
    \label{fig:appendix1}
\end{figure}

\end{document}